\pdfoutput=1

\documentclass[11pt]{article}

\usepackage[preprint]{acl}

\usepackage{times}
\usepackage{latexsym}
\usepackage[subtle]{savetrees}
\usepackage[T1]{fontenc}

\usepackage[utf8]{inputenc}

\usepackage{microtype}

\usepackage{inconsolata}

\usepackage{graphicx}
\usepackage{amsfonts}
\usepackage{amsmath}
\usepackage{bbm}
\usepackage{booktabs}
\usepackage{array}

\usepackage{wrapfig}
\usepackage{inconsolata}
\usepackage{xspace}
\usepackage{graphicx}
\usepackage{booktabs} 
\usepackage{marvosym}  
\usepackage{graphicx}
\usepackage{caption}
\usepackage{stfloats} 
\usepackage{cuted}
\usepackage{multirow}
\usepackage{tabularx}
\usepackage{subcaption}
\usepackage{flushend}

\usepackage{algorithm}
\usepackage{algpseudocode}

\newcommand\CONDITION[2]%
   {\begin{tabular}[t]{@{}l@{}}#1#2\end{tabular}}
\algdef{SE}[WHILE]{While}{EndWhile}[1]%
   {\algorithmicwhile\ \CONDITION{#1}{\ \algorithmicdo}}%
   {\algorithmicend\ \algorithmicwhile}
\algdef{SE}[FOR]{For}{EndFor}[1]%
   {\algorithmicfor\ \CONDITION{#1}{\ \algorithmicdo}}%
   {\algorithmicend\ \algorithmicfor}
\algdef{S}[FOR]{ForAll}[1]%
   {\algorithmicforall\ \CONDITION{#1}{\ \algorithmicdo}}
\algdef{SE}[REPEAT]{Repeat}{Until}%
   {\algorithmicrepeat}[1]%
   {\algorithmicuntil\ \CONDITION{#1}{}}
\algdef{SE}[IF]{If}{EndIf}[1]%
   {\algorithmicif\ \CONDITION{#1}{\ \algorithmicthen}}%
   {\algorithmicend\ \algorithmicif}
\algdef{C}[IF]{IF}{ElsIf}[1]%
   {\algorithmicelse\ \algorithmicif\ \CONDITION{#1}{\ \algorithmicthen}}
\makeatletter
\ifthenelse{\equal{\ALG@noend}{t}}%
   {\algtext*{EndWhile}%
    \algtext*{EndFor}%
    \algtext*{EndLoop}%
    \algtext*{EndIf}%
   }{}%
\makeatother

\usepackage{tocloft}
\usepackage{etoc}

\usepackage{adjustbox}
\usepackage{enumitem}

\usepackage[table]{xcolor}
\usepackage{lipsum}
\usepackage{titletoc}
\usepackage{tcolorbox}
\usepackage{arydshln}
\usepackage{comment}

\tcbuselibrary{breakable}

\newif\ifreview
\reviewtrue       
\reviewfalse     

\newcommand{\hh}[1]{%
\ifreview
{\textcolor{orange!90!black}{#1}}
\else
#1
\fi
}

\newcommand{\tcbtab}{\hspace*{4ex}}
\newcommand{\hyphentt}[1]{\texttt{\hyphenchar\font=\defaulthyphenchar #1}}

\algrenewcommand\algorithmicrequire{\textbf{Input:}}
\algrenewcommand\algorithmicensure{\textbf{Output:}}

\expandafter\def\expandafter\normalsize\expandafter{%
    \normalsize%
    \setlength\abovedisplayskip{0pt}%
    \setlength\belowdisplayskip{2pt}%
}
\title{Can LLMs Write Reliable Rubrics? \\ A Meta-Evaluation for Experiment Reproduction}

\author{
 \textbf{Hanhua Hong\textsuperscript{1,2}},
 \textbf{Yizhi Li\textsuperscript{3}},
 \textbf{Jiaoyan Chen\textsuperscript{1}},
\\
 \textbf{Luu Gia Huy\textsuperscript{5}},
 \textbf{Sophia Ananiadou\textsuperscript{1,4}},
 \textbf{Jung-jae Kim\textsuperscript{2}},
 \textbf{Chenghua Lin\textsuperscript{1}}
\\
 \textsuperscript{1}The University of Manchester, 
 \textsuperscript{2}Institute for Infocomm Research (I²R), A*STAR
 \\
 \textsuperscript{3}IQuest Research
 \textsuperscript{4}ELLIS Manchester, 
 \textsuperscript{5}University of Information Technology, VNU
 \\
 \texttt{hanhua.hong@postgrad.manchester.ac.uk}, \texttt{kim\_jung\_jae@a-star.edu.sg}
  \\
 \texttt{\{jiaoyan.chen,sophia.ananiadou,chenghua.lin\}@manchester.ac.uk}
}

\begin{document}

    \maketitle
\begin{abstract}
Rubric-based evaluation is a promising approach for assessing open-ended outputs from LLM-based research agents, particularly in paper reproduction, where direct paper-to-repository comparison is prone to hallucination. However, constructing paper-specific rubrics requires substantial expert effort, limiting the scalability of benchmarks such as PaperBench. In this work, we present, to our knowledge, the first systematic meta-evaluation of LLM-generated rubrics for paper reproduction. We reformulate rubrics into a checklist-style format and evaluate four generation settings across two backbone models. We meta-evaluate generated rubrics intrinsically by semantic similarity and extrinsically by score alignment with ground-truth rubrics. Our results show that the augmented settings substantially improves downstream evaluation alignment, with the strongest setting approaching the human baseline, while intrinsic gains are more modest. Further analyses reveal that LLM-generated rubrics are often overly fine-grained, biased toward high scores, and less adaptive to paper domains, highlighting both the affordances and limitations.
\end{abstract}

\begin{figure*}[htb]
    \centering
    \includegraphics[width=0.95\textwidth]{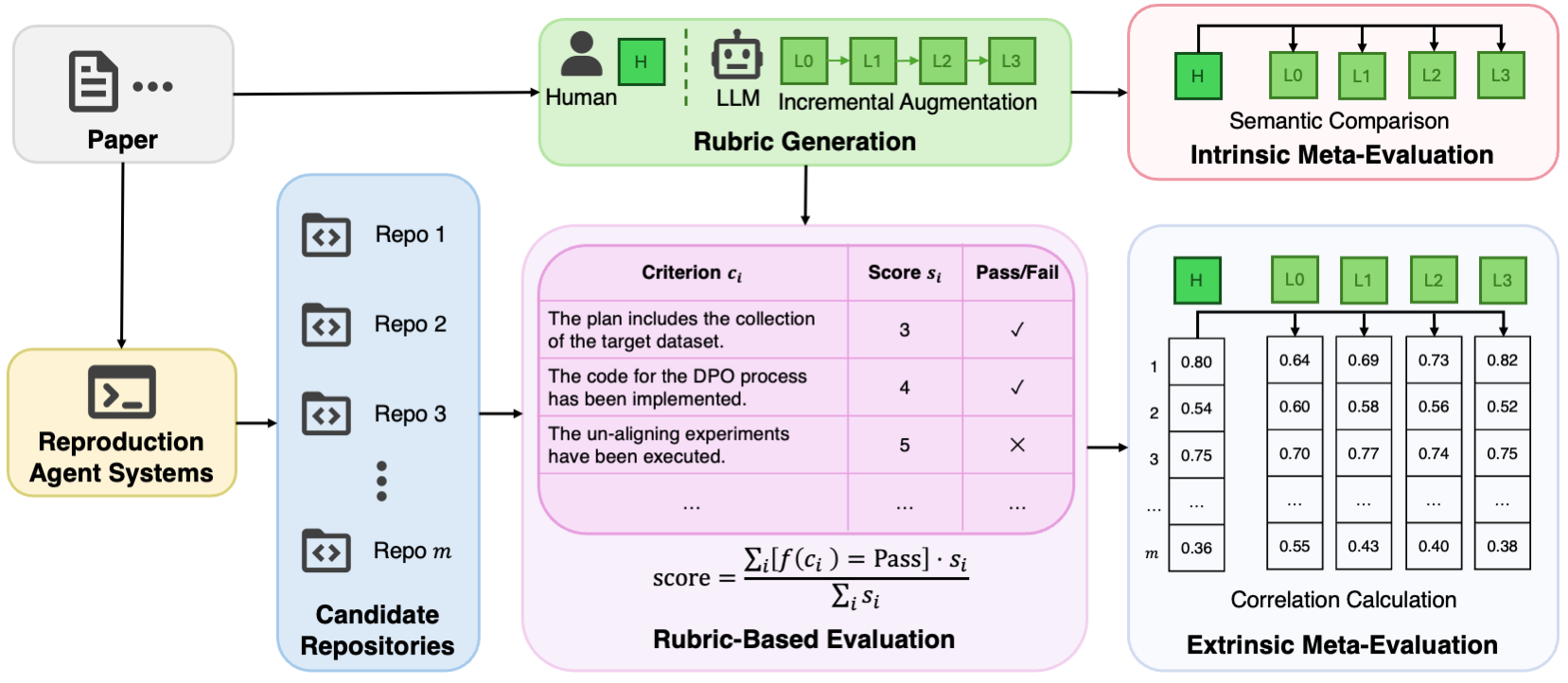}
    \caption{Overview of our meta-evaluation framework for LLM-generated rubrics in paper reproduction. Given a paper, reproduction agents generate candidate repositories, while LLMs and human experts produce checklist-style rubrics. We evaluate generated rubrics intrinsically by measuring their semantic similarity to human-authored rubrics, and extrinsically by comparing the repository scores they induce with PaperBench evaluations.}
    \label{fig:main_figure}
    \vspace{-1.5em}
\end{figure*}

\section{Introduction}

Recent advances in large language models (LLMs) have substantially improved capabilities in code generation, reasoning, and tool use, driving rapid progress in LLM-based agents for scientific research~\cite{zhao2025autoreproduceautomaticaiexperiment,yuan-etal-2025-dolphin,yamada2025aiscientistv2workshoplevelautomated,lyu2026evoscientistmultiagentevolvingai}. These agents are increasingly expected to interpret research papers, design experimental pipelines, and execute computational studies with minimal human supervision.

Within this workflow, reproducing the experiments in a paper as an executable code repository remains one of the most critical and challenging stages~\cite{zhu2025aiscientistsfailstrong}. The challenge is particularly severe in machine learning (ML), where publication pace far exceeds the release of artefacts, with only around 20\% of recent papers at top conferences providing complete and reproducible codebases~\cite{doi:10.1142/S0218194022500358,magnusson-etal-2023-reproducibility,seo2025paper2codeautomatingcodegeneration}. To address this bottleneck, a growing line of work focuses on automated paper-to-code generation, including PaperCoder~\cite{seo2025paper2codeautomatingcodegeneration}, AutoP2C~\cite{lin2025autop2cllmbasedagentframework}, AutoReproduce~\cite{zhao2025autoreproduceautomaticaiexperiment}, and HiRAS~\cite{hong2026hirashierarchicalmultiagentframework}.

However, reliably evaluating these systems is itself just as challenging, if not more. Direct paper-to-repository comparison, as adopted in the Paper2Code benchmark~\cite{seo2025paper2codeautomatingcodegeneration}, is vulnerable to hallucination, where evaluators incorrectly infer that described methods have been implemented~\cite{hong2026hirashierarchicalmultiagentframework}. In contrast, PaperBench~\cite{starace2025paperbenchevaluatingaisability} introduces \emph{rubric-based evaluation}: each paper is paired with a tree-structured rubric consisting of fine-grained, weighted pass/fail criteria spanning distinct reproduction stages, and candidate repositories are assessed independently against each criterion. Rubric-based evaluation has increasingly become a common strategy for evaluating open-ended LLM outputs across domains~\cite{wang2025profbenchmultidomainrubricsrequiring,arora2025healthbenchevaluatinglargelanguage,sharma2025researchrubricsbenchmarkpromptsrubrics}, and has been shown to yield more reliable judgements than holistic evaluation~\cite{zhou-etal-2026-rubricbench}.

Yet rubric construction has itself emerged as a major bottleneck. PaperBench rubrics are manually designed in close collaboration with original paper authors, requiring substantial human effort while still covering only 20 papers. The number of criteria per paper ranges from 94 to 2{,}551, reflecting the difficulty of standardising rubric design across diverse papers and annotators. Similar annotation costs arise in rubric benchmarks across other domains: HealthBench required contributions from 262 physicians~\cite{arora2025healthbenchevaluatinglargelanguage}, while ResearchRubrics consumed over 2{,}800 expert hours~\cite{sharma2025researchrubricsbenchmarkpromptsrubrics}. At the same time, LLMs are increasingly used both as judges operating over fixed rubrics~\cite{zheng2023judging,gu2025surveyllmasajudge} and as rubric generators for Reinforcement Learning reward modelling~\cite{gunjal2026rubrics}. Whether LLMs can also serve as reliable generators of rubrics for high-stakes paper-reproduction evaluation, while preserving alignment with human judgements, remains an open question.

\hh{In this work, we investigate this question by first formulating rubrics in a checklist-style representation, which is widely adopted in rubric-based evaluation~\cite{sharma2025researchrubricsbenchmarkpromptsrubrics,arora2025healthbenchevaluatinglargelanguage,wang2025profbenchmultidomainrubricsrequiring}.} We then study four incrementally augmented rubric-generation settings, ranging from direct prompting to in-context examples, agentic scaffolding, and distilled procedural skills. Finally, we meta-evaluate the generated rubrics along two complementary axes: intrinsic alignment with human-authored rubrics, measured by lexical and semantic similarity, and extrinsic alignment, measured by the correlation between repository-level scores induced by generated and ground-truth rubrics. To our knowledge, this is the first systematic meta-evaluation of model-generated rubrics for paper reproduction. \hh{An overview of our work is presented in Figure~\ref{fig:main_figure}.}

In summary, our contributions are three-fold:

\setlist{nolistsep}
\begin{itemize}

\item We present a controlled meta-evaluation of LLMs as rubric authors for ML paper reproduction on PaperBench, covering four incrementally augmented generation settings and two backbone models. The strongest setting achieves a Spearman correlation of $0.78$ in extrinsic alignment, approaching the human baseline of $0.83$.

\item We show that incremental augmentation substantially improves extrinsic alignment with human PaperBench evaluations, while intrinsic similarity improves more modestly and is mainly driven by increased recall, indicating that downstream evaluation quality depends more on rubric coverage and granularity than textual similarity.

\item We further characterise the limitations of LLM-generated rubrics through rubric statistics, score-distribution analysis, category-level analysis, and qualitative comparison. Our findings show that generated rubrics are often more fine-grained, code-centric, and less adaptive to topic-specific reproduction priorities than human-authored rubrics.

\end{itemize}

\section{Related Work}

\subsection{Agent-Based Automatic Research}

Recent advances in LLMs have enabled agent-based systems for automating scientific workflows, from full-cycle research systems such as AI Scientist~\cite{DBLP:journals/corr/abs-2408-06292,yamada2025aiscientistv2workshoplevelautomated}, DOLPHIN~\cite{yuan-etal-2025-dolphin}, and Zochi~\cite{zochi2025}, to stage-specific tools for idea generation~\cite{wang-etal-2024-scimon,li-etal-2025-chain-ideas}, rigour assessment~\cite{james-etal-2024-rigour}, and experiment reproduction. In the reproduction setting, systems such as PaperCoder~\cite{seo2025paper2codeautomatingcodegeneration}, AutoP2C~\cite{lin2025autop2cllmbasedagentframework}, AutoReproduce~\cite{zhao2025autoreproduceautomaticaiexperiment}, and HiRAS~\cite{hong2026hirashierarchicalmultiagentframework} aim to reconstruct codebases from research papers.

Alongside these systems, benchmarks such as MLS-Bench~\cite{lyu2026mlsbenchholisticrigorousassessment}, SciReplicateBench~\cite{xiang2025scireplicatebench}, and PaperBench~\cite{starace2025paperbenchevaluatingaisability} have been proposed to evaluate scientific agents. While these benchmarks improve evaluation reliability, they also expose a key bottleneck: \hh{they require fine-grained, expert criteria which are normally not available}. Our addresses this bottleneck by systematically investigating whether LLMs can generate paper-specific rubrics aligned with human-authored evaluations.

\subsection{Rubric-Based Evaluation}

Rubrics have recently gained attention as structured supervision signals that decompose complex goals into fine-grained criteria. In LLM post-training, instance-specific rubrics have been used as reward signals to extend Reinforcement Learning beyond verifiable domains such as mathematics and coding~\cite{gunjal2026rubrics}, while recent work explores scalable synthetic rubric generation for reward modelling and alignment~\cite{liu2026openrubricsscalablesyntheticrubric}. Rubrics also support structured evaluation by replacing holistic judgments with explicit criteria. In paper reproduction, PaperBench pairs each paper with a hierarchical rubric co-developed with original authors~\cite{starace2025paperbenchevaluatingaisability}, and similar protocols have been adopted in HealthBench, ProfBench, and ResearchRubrics~\cite{arora2025healthbenchevaluatinglargelanguage,wang2025profbenchmultidomainrubricsrequiring,sharma2025researchrubricsbenchmarkpromptsrubrics}.

Despite their value, rubrics remain expensive to construct and difficult to standardise. PaperBench covers only 20 papers, while HealthBench required rubrics from 262 physicians~\cite{starace2025paperbenchevaluatingaisability,arora2025healthbenchevaluatinglargelanguage}. This has motivated work on model-generated rubrics and rubric-guided benchmarks, such as RubricBench~\cite{zhou-etal-2026-rubricbench}. Our work studies this problem in paper reproduction, where rubric quality depends not only on textual plausibility but also on its ability to induce repository-level evaluations aligned with human-authored rubrics.

\subsection{Meta-Evaluation}

Meta-evaluation studies whether evaluation methods themselves are reliable and aligned with human judgments. This issue has become central as LLMs are increasingly used as automatic judges for open-ended generation. Prior studies such as G-Eval \citep{liu-etal-2023-g} show that prompting LLMs with explicit evaluation criteria can improve correlation with human judgments, while LLM-as-a-judge studies and surveys identify important reliability concerns, including verbosity bias, position bias, self-preference, and inconsistency~\cite{liu-etal-2023-g,zheng2023judging,gu2025surveyllmasajudge}.

For rubric-based evaluation, both the evaluator and the rubric itself are critical. PaperBench compares human and LLM evaluators under fixed human-authored rubrics, while RubricBench primarily assesses evaluators under different rubric sources in Reinforcement Learning reward modelling settings, with limited analysis of rubric generation itself~\cite{starace2025paperbenchevaluatingaisability,zhou-etal-2026-rubricbench}. In contrast, our work directly focuses on meta-evaluating rubric generation for paper reproduction, exploring incrementally augmented generation settings through both intrinsic and extrinsic meta-evaluation.

\section{Methodology}
\subsection{Problem Definition}
In this work, we represent rubrics in a checklist-style format. Each rubric item $r_i$ consists of a criterion $c_i$ and an associated importance score, $s_i$ ranging from 1 to 5, resulting in $r_i=(c_i,s_i)$. For a paper $\mathcal{P}$, the complete rubric set is defined as $R=\{r_1,r_2,\dots,r_n\}$.

To evaluate whether a repository faithfully reproduces the experiments in the paper, the evaluator examines each rubric criterion independently and assigns a binary judgement, denoted as $f: c_i \rightarrow \text{Pass/Fail}$. The final reproduction score of the paper is computed as the weighted sum of all passed criteria, normalised by the total rubric score:

\begin{equation}
    \mathrm{score}=\frac{\sum_i\left[f(c_i)=\mathrm{Pass}\right]\cdot s_i}{\sum_is_i},
\end{equation}
where $[\cdot]$ returns 1 if the condition is satisfied and 0 otherwise.

\begin{algorithm}[t]
\small
\caption{Algorithm for the Agentic Scaffold}
\label{alg:agent}
\begin{algorithmic}[1]
\Require Backbone Model $\mathrm{LLM}$; Initial Prompt $p_0$; Workspace $W=\{\mathcal{P},E\}$
\Ensure Ending Report $\mathrm{Report}$; Updated Workspace $W$;
    \State $\mathrm{Mem}\gets p_0$.
    \While{$\mathrm{True}$}
        \State $\mathrm{Reasoning,Action}\gets \mathrm{LLM}(\mathrm{Mem})$.
            \State $\mathrm{Mem}\gets\mathrm{Mem}\cup\{(\mathrm{Reasoning,Action})\}$.
        \If{$\mathrm{Action}\in \mathrm{SystemTools}$}
            \State $\mathrm{Result},W\gets\texttt{system.call(}\mathrm{Action}\texttt{)}$.
            \State $\mathrm{Mem}\gets\mathrm{Mem}\cup\{ \mathrm{Result}\}$.
        \ElsIf{$\mathrm{Action} = \texttt{"End}\texttt{"}$}
            \State \textbf{break}
        \EndIf
    \EndWhile
    \State $\mathrm{Report}\gets\mathrm{Mem}$
    \State \textbf{return} $\mathrm{Report},W$.
\end{algorithmic}
\end{algorithm}
\subsection{Rubric Generation}

We study rubric generation under four incrementally augmented generation settings designed to improve the quality and reliability of LLM-authored rubrics. Each setting builds on the previous one by introducing an additional form of augmentation.

\noindent\textbf{Direct Prompting.}~~In the base setting, the model is directly prompted with \hh{the target paper $\mathcal{P}$ and a manually designed instruction set $I$ that specifies the checklist-style rubric format and scoring standards. We denote the resulting prompt as $p_{DP}=(I,\mathcal{P})$.} The model then generates the complete rubric in a single pass.

\noindent\textbf{In-Context Examples.}~~Building on Direct Prompting, we additionally provide human-authored rubric examples $E$ from other papers as in-context demonstrations. These examples expose the model to the desired rubric structure, criterion granularity, and score allocation patterns, encouraging more consistent and human-aligned rubric generation. \hh{We denote the resulting prompt as $p_{IE}=(I,\mathcal{P},E)$.}

\noindent\textbf{Agentic Scaffold.}~~
The third setting introduces an agentic scaffold following the ReAct paradigm~\cite{yao2023react}. Rather than generating the entire rubric in a single response, the model iteratively reasons over the task, inspects intermediate outputs, and refines the generated criteria across multiple interaction steps. \hh{The agent starts with an initial prompt $p_0=(I)$ and operates in a workspace $W=\{\mathcal{P},E\}$, which initially contains the target paper and in-context examples. The final generated rubric will also be stored in this workspace.} The agent is equipped with system tools such as $\mathrm{Read}$ and $\mathrm{Write}$, enabling it to access relevant files and save intermediate or final outputs during rubric generation. \hh{The full procedure is provided in Algorithm~\ref{alg:agent}.}

\noindent\textbf{Distilled Skill.}~~Finally, we augment the generation process with a distilled procedural skill extracted from a stronger model. \hh{Instead of using the stronger model to generate rubrics for every target paper, which would incur substantial computational cost, we run it only once on a single paper under the Agentic Scaffold setting. After this run, the agent is further prompted to summarise its rubric-construction procedure, heuristics, and decision patterns into a general-purpose skill document $S$.} This distilled skill is then incorporated into the prompt of the weaker model and reused to guide rubric generation for all papers. \hh{This setting also follows the agentic procedure, with the initial prompt $p_0=(I,S)$ and workspace $W=\{\mathcal{P},E\}$.}

 
These four settings form a controlled augmentation hierarchy, enabling us to systematically analyse how progressively stronger forms of supervision and reasoning support affect the quality and evaluation behaviour of generated rubrics. The prompt details are provided in Appendix~\ref{app:prompts}.

\subsection{Meta-Evaluation Metrics}

We conduct meta-evaluation from two complementary perspectives. \textit{Extrinsic meta-evaluation} assesses whether generated rubrics induce repository-level evaluation outcomes consistent with those obtained using ground-truth rubrics. \textit{Intrinsic meta-evaluation} measures the \hh{lexical and semantic} alignment between model-generated and human-authored rubrics.

\noindent\textbf{Extrinsic Meta-Evaluation.}~~
For extrinsic evaluation, we measure the alignment between generated and ground-truth rubrics through their induced repository-level evaluation outcomes. For each paper, we collect multiple reproduction repositories and evaluate them with the same LLM-based evaluator under different rubric sets. To fit within the context limit, for each rubric criterion, the evaluator first retrieves the top-$K$ most relevant files from the repository and then assigns a binary pass/fail judgment based on the retrieved contents.

We quantify outcome alignment using both Spearman and Pearson correlation coefficients. Given $m$ repositories, let $\mathcal{D}=\{d_1,\dots,d_m\}$ denote the repository set for a paper. The repository-level score vectors induced by the generated rubric $R$ and the ground-truth rubric $\mathrm{GT}$ are defined as:
\begin{equation}
\begin{split}
    \mathbf{s}^{R}
    &=
    \langle
    \mathrm{Eval}(d_i, R)
    \rangle_{i=1}^{m}, \\
    \mathbf{s}^{\mathrm{GT}}
    &=
    \langle
    \mathrm{Eval}(d_i, \mathrm{GT})
    \rangle_{i=1}^{m}.
\end{split}
\end{equation}

We then compute, for example, the Spearman correlation between the two score vectors:
\begin{equation}
    \rho
    =
    \mathrm{Spearman}
    \left(
    \mathbf{s}^{R},
    \mathbf{s}^{\mathrm{GT}}
    \right).
\end{equation}

The final extrinsic alignment score is obtained by averaging the correlation coefficients across all evaluated papers in PaperBench.

\noindent\textbf{Intrinsic Meta-Evaluation.}~~
We evaluate intrinsic alignment at both lexical and semantic levels. At the lexical level, for each generated rubric item, we compute its maximum ROUGE-1 and ROUGE-L scores~\cite{lin-2004-rouge} against all ground-truth rubric items for the same paper, and then average the scores across papers.

At the semantic level, following~\citet{zhou-etal-2026-rubricbench}, we measure matching precision, recall, and F1 between the generated rubric set
$R=\{r_i\}_{i=1}^{|R|}$, where $r_i=(c_i,s_i)$, and the ground-truth rubric set
$\mathrm{GT}=\{\mathrm{gt}_j\}_{j=1}^{|\mathrm{GT}|}$, where $\mathrm{gt}_j=(c^{\mathrm{gt}}_j,s^{\mathrm{gt}}_j)$. Unlike~\citet{zhou-etal-2026-rubricbench}, which relies on LLM-based matching, we use embedding cosine similarity~\cite{reimers-gurevych-2019-sentence} to identify criterion matches, thereby reducing the risk of hallucinations:
\begin{equation}
\mathrm{match}(c_i,c^{\mathrm{gt}}_j)
=
[\cos(c_i,c^{\mathrm{gt}}_j) > \tau],
\end{equation}
where $\tau$ is a predefined similarity threshold.

\hh{A generated criterion is considered matched if it matches at least one ground-truth criterion}, while a ground-truth criterion is considered covered if it is matched by at least one generated criterion. Precision, recall, and F1 are computed as:
\begin{equation}
\mathrm{Precision}
=
\frac{
{\sum}_i
\left[
{\sum}_j
\mathrm{match}(c_i,c^{\mathrm{gt}}_j) \geq 1
\right]
}{|R|},
\end{equation}

\begin{equation}
\mathrm{Recall}
=
\frac{
\sum_{j}
\left[
\sum_{i}
\mathrm{match}(c_i,c^{\mathrm{gt}}_j) \geq 1
\right]
}{|\mathrm{GT}|},
\end{equation}

\begin{equation}
\mathrm{F1}
=
\frac{2 \cdot \mathrm{Precision} \cdot \mathrm{Recall}}
{\mathrm{Precision} + \mathrm{Recall}}.
\end{equation}

All semantic metrics are also computed per paper and then averaged across all 
evaluated papers in PaperBench.


\section{Experiments}

\begin{table*}[t]
    \centering
    \begin{adjustbox}{max width=0.95\textwidth}
    \small
    \begin{tabular}{lcccccc}
        \toprule
        \multirow{2}{*}{\textbf{Method}} & \multirow{2}{*}{\textbf{Backbone Model}} & \multicolumn{2}{c}{\textbf{Correlation Coefficients}} & \multicolumn{3}{c}{\textbf{Average Statistics}}\\
        \cmidrule(lr){3-4} \cmidrule(lr){5-7}
        && \textbf{Pearson} & \textbf{Spearman} & \textbf{avg. count} & \textbf{avg. len} & \textbf{avg. tokens}\\
        \midrule
        Human Annotation & - & 0.83 & 0.75 & 83.20 & 131.73 & 29.49 \\
        \midrule
        Direct Prompting & \multirow{4}{*}{\hyphentt{Claude-Sonnet}} & 0.62 & 0.58 & 35.95 & 185.41 & 45.05 \\
        ~\textit{+In-Context Examples} & & 0.70 & 0.65 & 56.30 & 158.80 & 38.42 \\
        ~\textit{+Agentic Scaffold} & & 0.73 & 0.67 & 57.85 & \textbf{140.30} & \textbf{33.97} \\
        ~\textit{+Distilled Skill} & & \textbf{0.78} & \textbf{0.70} & \textbf{77.95} & 171.85 & 40.87 \\
        \midrule
        Direct Prompting & \multirow{4}{*}{\hyphentt{GPT-5.4}}  & 0.55 & 0.55 & 50.15 & 190.30 & 37.75 \\
        ~\textit{+In-Context Examples} & & 0.56 & 0.55 & 61.80 & 165.65 & 33.43 \\
        ~\textit{+Agentic Scaffold} & & 0.60 & 0.63 & 57.60 & 156.15 & 32.05 \\
        ~\textit{+Distilled Skill} & & \textbf{0.74} & \textbf{0.70} & \textbf{87.15} & \textbf{138.70} & \textbf{29.58} \\
        \bottomrule
    \end{tabular}
    \end{adjustbox}
    \vspace{-0.5em}
    \caption{Extrinsic meta-evaluation results. For Spearman and Pearson correlations, bold values indicate the best performance of each model. For average rubric statistics, bold values indicate the closest match to the human baseline. Incremental augmentations generally improve alignment with human PaperBench evaluations and increase rubric count. The Distilled Skill setting achieves the strongest model-generated performance, approaching the Human Annotation baseline.}
    \label{tab:extrinsic_results}
    \vspace{-1.5em}
\end{table*}

\noindent\textbf{Models.}~~For rubric generation, we use \hyphentt{Claude-Sonnet} and \hyphentt{GPT-5.4} to generate rubrics for all settings included in the meta-evaluation. Skill distillation is conducted with \hyphentt{Claude-Opus}~\cite{singh2026openaigpt5card,claude2026opus4.6card,claude2026sonnet4.6card}. For evaluators, we use \hyphentt{ChatGPT-4o-mini} to assign scores to each repository according to the rubrics~\cite{openai2024gpt4ocard}. In addition, we use \hyphentt{Qwen3-Embedding-8B}~\cite{zhang2025qwen3embeddingadvancingtext} to calculate the cosine similarity for the intrinsic meta-evaluation.


\noindent\textbf{Baselines.}~~For comparison, we also manually re-annotated the PaperBench papers in checklist-style format to evaluate \hh{how human-authored rubrics} align with the ground-truth rubrics of PaperBench. This serves as a human baseline, which is denoted as Human Annotation in the main result tables. For the In-Context Examples setting, we add one set of our human-authored rubrics in to the prompt to instruct model to generate rubrics in the similar format. For the Agentic Scaffold setting, we use the \hyphentt{Claude Code} scaffold~\cite{claude2026claudecode}.

\noindent\textbf{Repositories.}~~For extrinsic meta-evaluation, we use five reproduction repositories generated by prior agent-based paper-reproduction systems. These repositories are produced under configurations of \hyphentt{PaperCoder} and \hyphentt{HiRAS}: \hyphentt{PaperCoder} is run with \hyphentt{DeepSeek-V3} and \hyphentt{Qwen3-Coder-480B}, while \hyphentt{HiRAS} is run with \hyphentt{Claude-Sonnet}, \hyphentt{DeepSeek-V3}, and \hyphentt{Qwen3-Coder-480B}~\cite{deepseekai2025deepseekv3technicalreport,yang2025qwen3technicalreport}.

\noindent\textbf{Environment.}~~All experiments are conducted using eight NVIDIA L40S GPUs. For repository evaluation, the evaluator retrieves the top-$K=10$ most relevant files for each rubric criterion. For semantic matching, we set the similarity threshold to $\tau=0.7$ in the main experiments. Our key findings are not sensitive to this choice: varying $\tau$ within a reasonable range preserves the same overall trends across augmentation levels to rubric generation. Additional threshold-sensitivity results are provided in Appendix~\ref{app:threshold_discussion}.

\begin{table*}[ht]
    \centering
    \small
    \begin{adjustbox}{max width=0.95\textwidth}
    \begin{tabular}{lcccccc}
    \toprule
    \multirow{2}{*}{\textbf{Method}} & \multirow{2}{*}{\textbf{Backbone Model}} & \multicolumn{3}{c}{\textbf{Cosine Similarity}} & \multicolumn{2}{c}{\textbf{avg. Max ROUGE}}\\
    \cmidrule(lr){3-5} \cmidrule(lr){6-7}
    && \textbf{Precision} & \textbf{Recall} & \textbf{F1} & \textbf{ROUGE-1} & \textbf{ROUGE-L}\\
    \midrule
    Human Annotation & - & 0.79 & 0.62 & 0.69 & 0.46 & 0.39 \\
    \midrule
    Direct Prompting & \multirow{4}{*}{\hyphentt{Claude-Sonnet}} & \textbf{0.76} & 0.58 & 0.63 & 0.38 & 0.29 \\
    ~\textit{+In-Context Examples} & & 0.74 & 0.60 & 0.65 & 0.39 & 0.31 \\
    ~\textit{+Agentic Scaffold} & & 0.71 & 0.61 & 0.64 & 0.39 & 0.31 \\
    ~\textit{+Distilled Skill} & & 0.74 & \textbf{0.65} & \textbf{0.68} & \textbf{0.41} & \textbf{0.32} \\
    \midrule
    Direct Prompting & \multirow{4}{*}{\hyphentt{GPT-5.4}} & 0.69 & 0.49 & 0.55 & 0.35 & 0.27 \\
    ~\textit{+In-Context Examples} & & 0.67 & 0.55 & 0.59 & 0.37 & 0.29 \\
    ~\textit{+Agentic Scaffold} & & \textbf{0.72} & 0.56 & \textbf{0.62} & 0.38 & 0.30 \\
    ~\textit{+Distilled Skill} & & 0.67 & \textbf{0.61} & \textbf{0.62} & \textbf{0.39} & \textbf{0.31} \\
    \bottomrule
\end{tabular}
    \end{adjustbox}
    \vspace{-0.5em}
    \caption{Intrinsic meta-evaluation results. Bold values denote the best performance for each model. Incremental augmentations generally improve lexical overlap and semantic matching with human-authored rubrics. The gains are most evident in cosine-similarity recall, F1, and ROUGE scores, indicating improved coverage of human-specified criteria, while precision generally fluctuates across settings. 
    }
    \label{tab:intrinsic_results}
    \vspace{-1.5em}
\end{table*}

\subsection{Extrinsic Meta-Evaluation Results}

Table~\ref{tab:extrinsic_results} presents the extrinsic meta-evaluation results under incrementally augmented rubric-generation settings. Overall, each augmentation step improves the alignment between model-generated and human-authored rubrics across both model families, validating the effectiveness of the proposed augmentation hierarchy. In particular, the full Distilled Skill setting achieves the strongest model-generated performance for both \hyphentt{Claude-Sonnet} and \hyphentt{GPT-5.4}, indicating that the proposed augmentations generalise across different backbone models.

For \hyphentt{Claude-Sonnet}, Direct Prompting yields substantially lower alignment than the human-annotation baseline, with a significant drop of approximately $0.2$ in both Spearman and Pearson correlations. As in-context examples, agentic scaffolding, and distilled procedural skill are introduced, the correlations improve monotonically, demonstrating that each component contributes to stronger repository-level evaluation alignment. Compared with Direct Prompting, the Distilled Skill setting improves Spearman correlation by approximately $26\%$ and Pearson correlation by approximately $21\%$, bringing the induced evaluation outcomes close to those obtained with human-authored rubrics. A similar pattern is observed for \hyphentt{GPT-5.4}. Although direct prompting starts from weaker alignment, performance increases as the augmentation level grows and reaches its best result under the Distilled Skill setting. Relative to direct prompting, Distilled Skill improves Spearman correlation by approximately $35\%$ and Pearson correlation by approximately $27\%$, yielding near-human performance and matching the best model-generated Pearson correlation. 
The initial drop of Direct Prompting is consistent with prior findings that naive LLM involvement in rubric construction can substantially degrade evaluation quality~\cite{sharma2025researchrubricsbenchmarkpromptsrubrics}. Nevertheless, our results further show that proposed augmentations can substantially improve rubric quality, enabling current LLMs to induce repository-level evaluation outcomes that approach those obtained with human-authored rubrics.

We also report rubric statistics in Table~\ref{tab:extrinsic_results}. Incremental augmentations also substantially increase rubric quantities, with the average number of criteria rising from $35.95$ to $77.95$ for \hyphentt{Claude-Sonnet} and from $50.15$ to $87.15$ for \hyphentt{GPT-5.4}. These counts approach, or even slightly exceed, the human average of $83.20$, indicating that stronger generation settings produce more comprehensive rubric sets. However, model-generated criteria remain generally longer than human-authored ones, reflecting a length bias in LLM generation~\cite{hu-etal-2025-explaining,bu-etal-2025-beyond}. Human annotations tend to aggregate implementation details into concise high-level criteria, whereas model-generated rubrics more often decompose them into fine-grained operational steps. This difference is further illustrated with the case study in \S\ref{sec:case_study}.

\begin{figure*}[h]
    \centering
    \begin{subfigure}[h]{0.45\textwidth}
        \centering 
        \includegraphics[width=\textwidth]{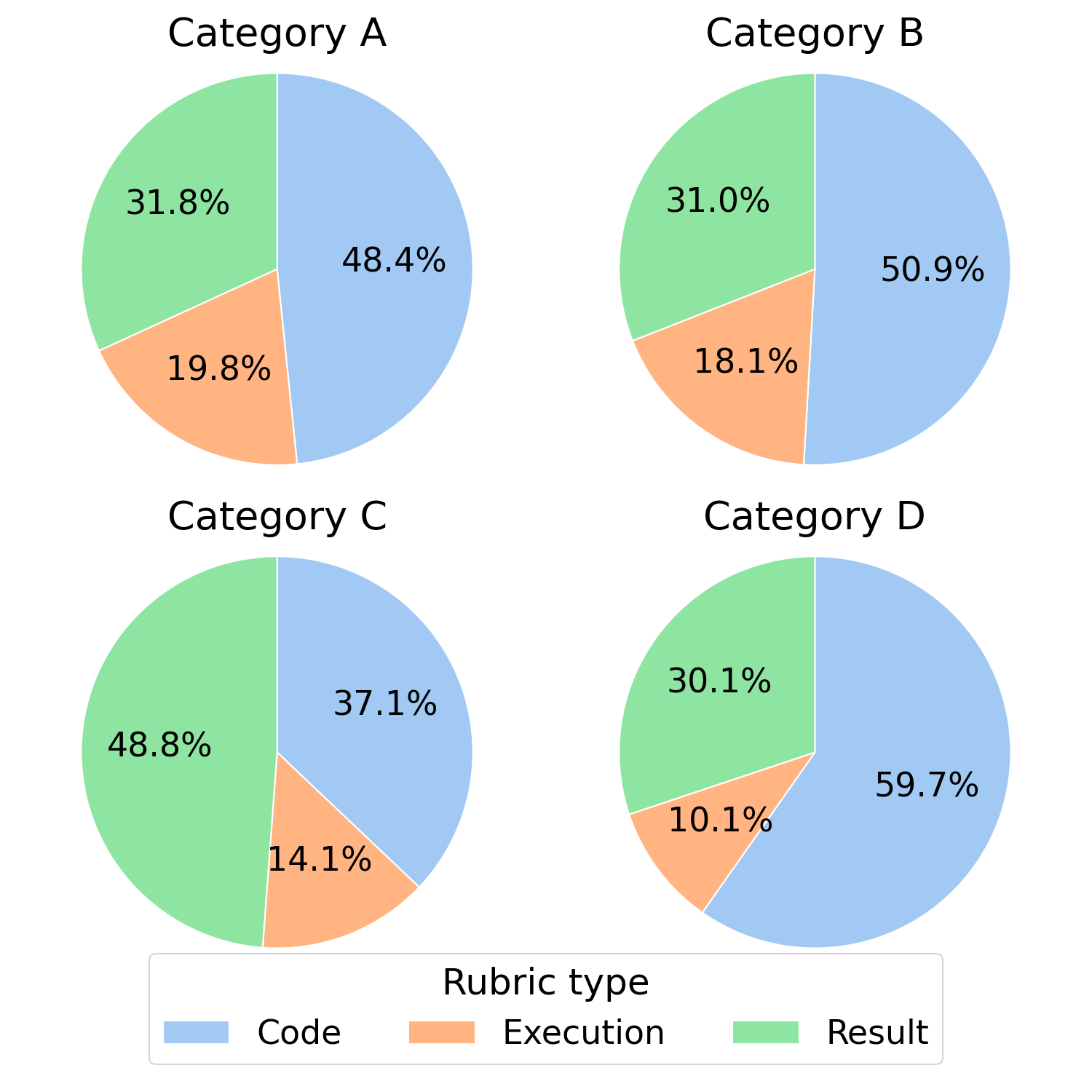}
        \caption{PaperBench Rubrics}
        \label{fig:direct_prompting}
    \end{subfigure}
    \begin{subfigure}[h]{0.45\textwidth}
        \centering 
        \includegraphics[width=\textwidth]{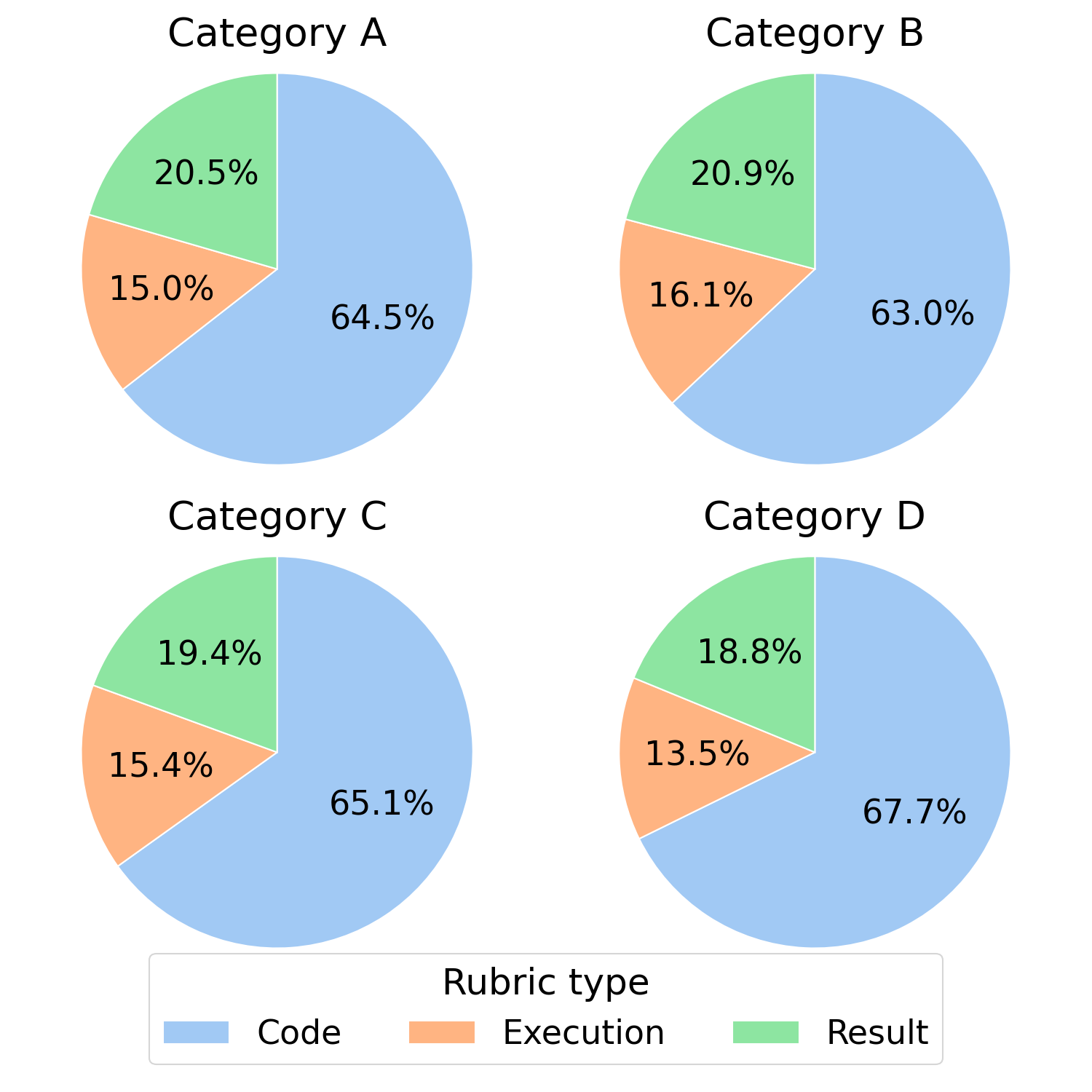}
        \caption{Model-Generated Rubrics}
        \label{fig:agent_scaffold}
    \end{subfigure}
    \vspace{-0.5em}
    \caption{Score proportions by rubric type across paper categories. Compared with the ground-truth PaperBench rubrics, model-generated rubrics from \hyphentt{Claude-Sonnet} under the Distilled Skill setting show more uniform distributions and assign consistently higher weight to \textit{Code Development} rubrics, indicating a code-centric bias and weaker adaptation to topic-specific reproduction 
    }
    \label{fig:category_analysis}
    \vspace{-1.5em}
\end{figure*}

\subsection{Intrinsic Meta-Evaluation Results}

Table~\ref{tab:intrinsic_results} presents the intrinsic meta-evaluation results. Overall, stronger augmentation improves the intrinsic alignment between model-generated and human-authored rubrics. Both ROUGE-1 and ROUGE-L increase consistently from Direct Prompting to the Distilled Skill setting for both models, indicating that augmented generation produces rubric items with greater lexical overlap with human-authored criteria. The cosine-similarity F1 score also generally improves across augmentation levels, mainly driven by increased recall, \hh{showing that stronger augmentation settings primarily improve coverage of ground-truth rubric criteria.}

For \hyphentt{Claude-Sonnet}, the Distilled Skill setting improves cosine-similarity F1 by approximately $8\%$ relative to Direct Prompting, while ROUGE-1 and ROUGE-L improve by approximately $8\%$ and $10\%$, respectively. For \hyphentt{GPT-5.4}, the gains are more pronounced: Distilled Skill improves cosine-similarity F1 by approximately $13\%$, ROUGE-1 by approximately $11\%$, and ROUGE-L by approximately $15\%$ over direct prompting. These improvements suggest that incremental augmentations not only enhance downstream evaluation alignment, but also make generated rubrics intrinsically closer to human-authored rubrics.

At the same time, the intrinsic gains are less pronounced than the extrinsic improvements. Precision fluctuates across settings but remains close to the human baseline. This may imply that models can already generate semantically valid criteria. However, taken together with lower recall and longer rubrics in Table~\ref{tab:extrinsic_results}, this pattern indicates a tendency toward overly fine-grained criteria, which limit models' holistic coverage of the ground-truth rubric. The main improvement in cosine-similarity matching comes from recall,  which further suggests that augmentations help mitigate this granularity issue by improving the coverage of reproduction criteria. Overall, the intrinsic results show that augmentation improves textual alignment and semantic coverage, while the remaining gap to human annotations reflects differences in coverage and granularity.

\section{Further Analysis}



\subsection{Paper Category Analysis}

To examine topic-level differences, we group the PaperBench papers into four categories: (A) Language Model Adaptation and Alignment (5 papers); (B) Vision and Vision-Language Models (5 papers); (C) Reinforcement Learning (4 papers), and (D) Probabilistic and Mathematical ML (6 papers). The detailed categorisation is provided in Appendix~\ref{app:paper_category}.

PaperBench divides rubric criteria into three types: \textit{Code Development}, \textit{Execution}, and \textit{Result Match}. We follow this taxonomy and compute the proportion of total score assigned to each type within each category. Figure~\ref{fig:category_analysis} compares the ground-truth PaperBench rubrics with those generated by \hyphentt{Claude-Sonnet} under the Distilled Skill setting.

The ground-truth rubrics show clear category-dependent variation. In particular, \textit{Result Match} accounts for 48.8\% of the score in Reinforcement Learning papers, indicating that human rubrics place greater emphasis on empirical outcomes in the Reinforcement Learning domain, where the generalisation under different circumstances is critical. In contrast, model-generated rubrics are much more uniform across categories: \textit{Code Development} consistently receives over 60\% of the total score, while \textit{Result Match} remains below 21\%. This reveals a systematic code-centric bias in model-generated rubrics. After further inspection, the bias can be attributed to that the model tends to decompose implementation procedures into fine-grained steps and assign them substantial weights to even the finest granularity. Overall, these results show that LLM-generated rubrics are structurally consistent but less adaptive to topic-specific reproduction priorities.

\begin{table*}[t]
    \centering
    \small
    \begin{adjustbox}{max width=0.95\textwidth}
        \begin{tabular}{p{0.9\linewidth}c}
            \toprule
            \textbf{Criterion} & \textbf{Score} \\
            \midrule

            \multicolumn{2}{l}{\textit{Human-authored rubrics}} \\

            \midrule

            The collection and processing of the Wikitext-2 dataset as prompts are planned. & 2 \\

            The construction of a pairwise toxicity dataset using PPLM is planned. & 2 \\

            The DPO process for GPT2 and Llama2 with the pairwise toxicity dataset is planned. & 4 \\

            The code for the construction of a pairwise toxicity dataset using PPLM has been implemented. & 3 \\

            The code for the DPO process for GPT2 and Llama2 with the pairwise toxicity dataset has been implemented. & 5 \\

            The hyperparameters for DPO and PPLM have been set as in Appendix E. & 1 \\

            The un-aligning DPO experiment on GPT2-DPO matches Table 4: toxicity, perplexity, and F1 increase after scaling up toxic key vectors. & 5 \\

            \midrule
            
            \multicolumn{2}{l}{\textit{Model-generated rubrics}} \\

            \midrule

            The plan includes building 24,576 pairwise preference samples using Wikitext-2 prompts: greedy GPT2 sampling for positive continuations and PPLM for negative continuations. & 3 \\

            The plan includes training DPO on GPT2-medium and Llama2-7b with learning rate $10^{-6}$, batch size 4, RMSProp optimizer, $\beta=0.1$, and validation patience 10. & 3 \\

            The code implements the DPO loss, where positive and negative continuation likelihood ratios are compared against the reference model. & 5 \\

            The code trains DPO on GPT2-medium using the pairwise dataset with the specified hyperparameters. & 5 \\

            The code implements PPLM with all specified hyperparameters, including step size, top-$k$, iterations, GM scale, and KL scale. & 3 \\

            The un-alignment experiment on GPT2-DPO by scaling the top 7 toxic key vectors by $10\times$ has been implemented. & 4 \\

            The execution outcomes match Table 4: scaling toxic key vectors reverts toxicity while preserving PPL and F1. & 4 \\

            \bottomrule

        \end{tabular}
        \end{adjustbox}
    \vspace{-0.5em}
    \caption{Qualitative comparison between human-authored and model-generated rubrics for the same DPO un-alignment experiment. The model-generated rubrics are extracted from the Distilled Skill setting with \hyphentt{Claude-Sonnet}. They are more fine-grained and assign relatively high scores to auxiliary details, whereas the human-authored rubrics use general, concise criteria with more discriminative score allocation.}
    \label{tab:qualitative_analysis}
    \vspace{-1.5em}
\end{table*}

\subsection{Score Distribution Analysis}

Figure~\ref{fig:distribution_analysis} in Appendix~\ref{app:score_distribution} illustrates the importance-score distributions under different rubric-generation settings with \hyphentt{Claude-Sonnet}. Direct Prompting produces a noticeably skewed distribution, with scores concentrated at higher importance levels and limited coverage of lower-scored criteria, suggesting that naive prompting tends to overestimate criterion importance. As in-context examples, agentic scaffolding, and the distilled skill are introduced, the distributions become bell-shaped and more balanced across the $1$--$5$ range, suggesting improved score calibration. Nevertheless, all model-generated settings still show a mild bias toward higher scores. Overall, stronger procedural guidance improves importance-score calibration, but score overestimation remains a persistent limitation.

\subsection{Qualitative Analysis}
\label{sec:case_study}
To further characterise the generated rubrics, we inspect the outputs of the best-performing Distilled Skill setting on \hyphentt{Claude-Sonnet} and compare them with human-authored rubrics in Table~\ref{tab:qualitative_analysis}. The table presents matched rubric segments for the same DPO un-alignment experiment from \textit{A Mechanistic Understanding of Alignment Algorithms: A Case Study on DPO and Toxicity}~\cite{10.5555/3692070.3693122}, one of the papers included in PaperBench. We select this example because both rubrics target the same experiment, yet exhibit substantial differences in criterion granularity, score assignment, and overall rubric construction.

 The model-generated rubric describes the experiment at a lower operational level, decomposing a single reproduction component into multiple fine-grained checks. It also explicitly specifies implementation details such as the number of samples in the dataset, and detailed hyperparameter values. In contrast, the human-authored rubric uses more concise, higher-level criteria that focus on core reproduction milestones, such as constructing the pairwise toxicity dataset, implementing DPO, and reproducing the un-alignment results. Peripheral details, such as hyperparameter settings, are aggregated into a separate low-weight criterion. \hh{In addition, the model-generated rubrics are assigned relatively higher scores than their human-authored counterpart. This example reinforces our previous findings that model-generated rubrics are generally longer, more fine-grained, and more likely to assign high scores to coding-level details than human-authored rubrics.}



\section{Conclusion}

In this work, we studied whether LLMs can serve as reliable rubric authors for scalable and high-quality paper-reproduction evaluation. We manually re-annotated rubrics for PaperBench into a checklist-style format as a baseline and evaluated LLM-generated rubrics under four progressively augmented settings, using both intrinsic textual alignment and extrinsic repository-level alignment with human PaperBench judgments. Our results show that proposed augmentations substantially improves the downstream utility of generated rubrics, with the best setting achieving the strongest model-generated performance and approaching the human baseline. Intrinsic improvements are more modest and mainly driven by recall, suggesting that augmentation primarily expands coverage of ground-truth rubrics. Further analyses show that model-generated rubrics remain more verbose, fine-grained, and code-centric than human-authored rubrics, with less paper topic sensitivity and calibrated weighting.

\section*{Limitations}

Our work focuses on the reproduction of machine learning papers, using the currently prevalent PaperBench benchmark as the testbed. Therefore, the generalisation of our findings to other scientific domains remains uncertain, as experimental workflows and reproduction requirements may vary substantially across fields. The evaluation scale is also limited by the size of PaperBench and the availability of models and reproduction repositories. In addition, rubric quality may still vary with different prompt design, in-context examples, and the implementation details of the agentic scaffold.

\section*{Ethical Considerations}

All data used in this work are collected from publicly available sources. Our goal is to support more reliable evaluation and reproduction of prior research, rather than to replace the creative, critical, and interpretive roles of human researchers. The proposed rubric-generation framework should therefore be viewed as an assistive tool for scalable evaluation, with human oversight remaining important in high-stakes scientific assessment. All human annotations and benchmark reformulations were conducted by research students with sufficient experience.

\bibliography{custom}


\appendix

\onecolumn
\section{Prompt Details}
\label{app:prompts}

\begin{figure}[h]
\centering
\tcbset{
    colback=gray!5!white,
    width=0.95\columnwidth,
    fontupper=\small,
    left=1pt,
    right=1pt,
    valign=center,
    before=\vspace{0pt},
    after=\vspace{0pt}
}
\begin{tcolorbox}[]
You are a serious examiner of experiment reproduction. 
Now you are given a paper. You task is to read the entire paper and then give a set of rubrics to determine whether the paper is reproduced correctly.
Please make sure you read and understand these instructions carefully.

\# Paper:
\{paper\_content\}

\# Instructions:

1. Please read the paper carefully and then give a set of rubrics to determine whether the paper is reproduced correctly.

2. You must focus on the core contents about methodology and experiment settings of the paper, and the criteria in your rubric should be very critical and specific.

3. The rubrics should be as comprehensive as possible, making sure it can reflect all the implementation details of the paper.

4. Your final target is to evaluate the reproduction of a codebase according to the paper. Thus, your rubrics should be able to be validated with the codebase.

5. Each rubric should contain a single criteria and a score ranged from 1 to 5 according to its importance in the reproduction.

6. Do not output anything else except the JSON rubrics.

\# Score Standard:

The score you assign to each criteria should be based on the following standard:

1 Not Important: This criteria only reflects a very minor detail of the paper, or is not closely related to the code implementation. The reproduction will still be executable even if this criteria is not met. (e.g. mentioning some specific hyperparameters in plans)

2 Less Important: This criteria is generally less important in terms of implementation, but the reproduction will be significantly affected if this criteria is not met, especially in the early stages of the reproduction. (e.g. reading the whole paper and understanding the overall methodology)

3 Important: This criteria is important during the reproduction, greatly influence the whole reproduction process and the final result. The final outcome will be significantly affected if this criteria is not met. (e.g. planning the repository architecture or each of the experiment settings in detail)

4 Very Important: This criteria describes a key step, component, or function in the reproduction according to the paper. It is an essential part of the implementation, the repository would not be executable if this criteria is not met. (e.g. coding part of a component or a useful cuntion, implementing a specific experiment setting, etc.)

5 Critical: This criteria is one of the most important standard to measure whether a reproduction is successful. The reproduction will not be considered complete if this requirement is not met or there are missing points for this criteria. (e.g. coding the key method from the paper or a critical component for the whole reproduction to be executable, running the main experiments, etc.)

\# Rubrics Design Steps:

1. Read the full paper: You need to first read the full paper carefully and understand the overall methodology and experiment settings. Do not miss even a single detail, and make sure you have a clear understanding of how the paper is implemented.

2. Analyse the related sections: Based on your understanding of the paper, analyse the related sections that are closely related to the reproduction. For example, the methodology section, the experiment settings section, the results section, the discussion section, etc.

3. Extract key process: For each experiment mentioned in the paper, extract the key components, algorithm, or function that is essential for the reproduction. They must be constructed in the codebase, and multiple rubrics may be introduced to examine each step of the implementation of these key processes.

4. Articulate the criteria: You will be required to articulate the criteria for all detailed steps of the reproduction. Please make sure your rubrics cover all stages and details of the reproduction, and each criteria should be specific and clear, which is able to evaluate the reproduction with the codebase.

5. Assign the score: Based on the importance of each criteria, assign the score to each criteria. Please refer to the Score Standard section for the detailed score standard.

6. Output the rubrics: You must output a valid JSON string of all the rubrics. Do not output anything else except the JSON rubrics.

\# Format:

You should output the rubrics in the following JSON format. 
Remember to add ```json and ``` at the beginning and end of the JSON:

```json\\
{[}\\
\tcbtab\{\\
\tcbtab\tcbtab"criteria": "...",\\
\tcbtab\tcbtab"score": "..."\\
\tcbtab\}\\
{]}\\
```

DO NOT output anything else except the JSON rubrics.

Your output:
\end{tcolorbox}
\caption{The basic prompt template for rubric generation in the Direct Prompting setting.}
\label{fig:prompt_template}
\end{figure}
\clearpage

\begin{figure}[h]
\centering
\tcbset{
    colback=gray!5!white,
    width=0.95\columnwidth,
    fontupper=\small,
    left=1pt,
    right=1pt,
    valign=center,
    before=\vspace{0pt},
    after=\vspace{0pt}
}
\begin{tcolorbox}[]
You are a serious examiner of experiment reproduction. 
Now you are given a paper. You task is to read the entire paper and then give a set of rubrics to determine whether the paper is reproduced correctly.
Please make sure you read and understand these instructions carefully.

\# Paper:
\{paper\_content\}

\# Instructions:

1. Please read the paper carefully and then give a set of rubrics to determine whether the paper is reproduced correctly.

2. You must focus on the core contents about methodology and experiment settings of the paper, and the criteria in your rubric should be very critical and specific.

3. The rubrics should be as comprehensive as possible, making sure it can reflect all the implementation details of the paper.

4. Your final target is to evaluate the reproduction of a codebase according to the paper. Thus, your rubrics should be able to be validated with the codebase.

5. Each rubric should contain a single criteria and a score ranged from 1 to 5 according to its importance in the reproduction.

6. Do not output anything else except the JSON rubrics.

\# Score Standard:

The score you assign to each criteria should be based on the following standard:

1 Not Important: This criteria only reflects a very minor detail of the paper, or is not closely related to the code implementation. The reproduction will still be executable even if this criteria is not met. (e.g. mentioning some specific hyperparameters in plans)

2 Less Important: This criteria is generally less important in terms of implementation, but the reproduction will be significantly affected if this criteria is not met, especially in the early stages of the reproduction. (e.g. reading the whole paper and understanding the overall methodology)

3 Important: This criteria is important during the reproduction, greatly influence the whole reproduction process and the final result. The final outcome will be significantly affected if this criteria is not met. (e.g. planning the repository architecture or each of the experiment settings in detail)

4 Very Important: This criteria describes a key step, component, or function in the reproduction according to the paper. It is an essential part of the implementation, the repository would not be executable if this criteria is not met. (e.g. coding part of a component or a useful cuntion, implementing a specific experiment setting, etc.)

5 Critical: This criteria is one of the most important standard to measure whether a reproduction is successful. The reproduction will not be considered complete if this requirement is not met or there are missing points for this criteria. (e.g. coding the key method from the paper or a critical component for the whole reproduction to be executable, running the main experiments, etc.)

\# Rubrics Design Steps:

1. Read the full paper: You need to first read the full paper carefully and understand the overall methodology and experiment settings. Do not miss even a single detail, and make sure you have a clear understanding of how the paper is implemented.

2. Analyse the related sections: Based on your understanding of the paper, analyse the related sections that are closely related to the reproduction. For example, the methodology section, the experiment settings section, the results section, the discussion section, etc.

3. Extract key process: For each experiment mentioned in the paper, extract the key components, algorithm, or function that is essential for the reproduction. They must be constructed in the codebase, and multiple rubrics may be introduced to examine each step of the implementation of these key processes.

4. Articulate the criteria: You will be required to articulate the criteria for all detailed steps of the reproduction. Please make sure your rubrics cover all stages and details of the reproduction, and each criteria should be specific and clear, which is able to evaluate the reproduction with the codebase.

5. Assign the score: Based on the importance of each criteria, assign the score to each criteria. Please refer to the Score Standard section for the detailed score standard.

6. Output the rubrics: You must output a valid JSON string of all the rubrics. Do not output anything else except the JSON rubrics.

\# Format:

Here is a sample rubrics file, examples.json, is provided as reference from another paper. 
Read through the examples carefully to understand how they are structured and designed.
You should strictly follow the format defined in examples.json. 

\# Examples
\{examples\}

Remember to add ```json and ``` at the beginning and end of the JSON.
DO NOT output anything else except the JSON rubrics.

Your output:
\end{tcolorbox}
\caption{The prompt template with examples in the In-Context Examples setting.}
\label{fig:in-context_prompt}
\end{figure}
\clearpage

\begin{figure}[h]
\centering
\tcbset{
    colback=gray!5!white,
    width=0.95\columnwidth,
    fontupper=\small,
    left=1pt,
    right=1pt,
    valign=center,
    before=\vspace{0pt},
    after=\vspace{0pt}
}
\begin{tcolorbox}[]
You are a serious examiner of experiment reproduction. 
Now you are given a paper. You task is to read the entire paper and then give a set of rubrics to determine whether the paper is reproduced correctly.
Please make sure you read and understand these instructions carefully.

Your working directory is \{working\_directory\}.
The paper is under the working directory and named \{file\_name\}.
You should read the paper carefully and understand its content comprehensively.

\# Instructions:

1. Please read the paper carefully and then give a set of rubrics to determine whether the paper is reproduced correctly.

2. You must focus on the core contents about methodology and experiment settings of the paper, and the criteria in your rubric should be very critical and specific. This means that not all sections from the papers should be designed with rubrics. Only the rubrics related to important sections such as preliminary experiments, methodology, experiment setup, and results should be designed.

3. The rubrics should be as comprehensive as possible, making sure it can reflect all the implementation details of the paper.

4. Your final target is to evaluate the reproduction of a codebase according to the paper. Thus, your rubrics should be able to be validated with the codebase.

5. Each rubric should contain a single criteria and a score ranged from 1 to 5 according to its importance in the reproduction.

6. Do not output anything else except the JSON rubrics.

\# Score Standard:

The score you assign to each criteria should be based on the following standard:

1 Not Important: This criteria only reflects a very minor detail of the paper, or is not closely related to the code implementation. The reproduction will still be executable even if this criteria is not met. (e.g. mentioning some specific hyperparameters in plans)

2 Less Important: This criteria is generally less important in terms of implementation, but the reproduction will be significantly affected if this criteria is not met, especially in the early stages of the reproduction. (e.g. reading the whole paper and understanding the overall methodology)

3 Important: This criteria is important during the reproduction, greatly influence the whole reproduction process and the final result. The final outcome will be significantly affected if this criteria is not met. (e.g. planning the repository architecture or each of the experiment settings in detail)

4 Very Important: This criteria describes a key step, component, or function in the reproduction according to the paper. It is an essential part of the implementation, the repository would not be executable if this criteria is not met. (e.g. coding part of a component or a useful cuntion, implementing a specific experiment setting, etc.)

5 Critical: This criteria is one of the most important standard to measure whether a reproduction is successful. The reproduction will not be considered complete if this requirement is not met or there are missing points for this criteria. (e.g. coding the key method from the paper or a critical component for the whole reproduction to be executable, running the main experiments, etc.)

\# Rubrics Design Steps:

1. Read the full paper: You need to first read the full paper carefully and understand the overall methodology and experiment settings. Do not miss even a single detail, and make sure you have a clear understanding of how the paper is implemented.

2. Analyse the related sections: Based on your understanding of the paper, analyse the related sections that are closely related to the reproduction. For example, the methodology section, the experiment settings section, the results section, the discussion section, etc.

3. Extract key process: For each experiment mentioned in the paper, extract the key components, algorithm, or function that is essential for the reproduction. They must be constructed in the codebase, and multiple rubrics may be introduced to examine each step of the implementation of these key processes.

4. Articulate the criteria: You will be required to articulate the criteria for all detailed steps of the reproduction. Please make sure your rubrics cover all stages and details of the reproduction, and each criteria should be specific and clear, which is able to evaluate the reproduction with the codebase.

5. Assign the score: Based on the importance of each criteria, assign the score to each criteria. Please refer to the Score Standard section for the detailed score standard.

6. Output the rubrics: You must output a valid JSON string of all the rubrics. Do not output anything else except the JSON rubrics. 

\# Examples

A sample rubrics file, examples.json, is provided as reference from another paper.
Read through the examples carefully to understand how they are structured and designed.
Once you have a clear understanding, generate the rubrics for the current paper (\{file\_name\}), strictly following the format defined in examples.json. 

Output the result to \{output\_path\}.
\end{tcolorbox}
\caption{The initial prompt for agents.}
\label{fig:in-context_prompt}
\end{figure}
\clearpage

\section{Score Distribution}
\label{app:score_distribution}
\vspace{2.5cm}
\begin{figure}[h]
    \centering
    \begin{subfigure}[h]{0.45\textwidth}
        \centering 
        \includegraphics[width=\textwidth]{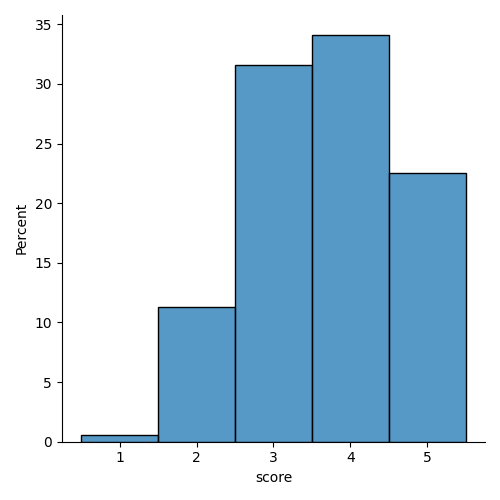}
        \caption{Direct Prompting}
        \label{fig:direct_prompting}
    \end{subfigure}
    \begin{subfigure}[h]{0.45\textwidth}
        \centering 
        \includegraphics[width=\textwidth]{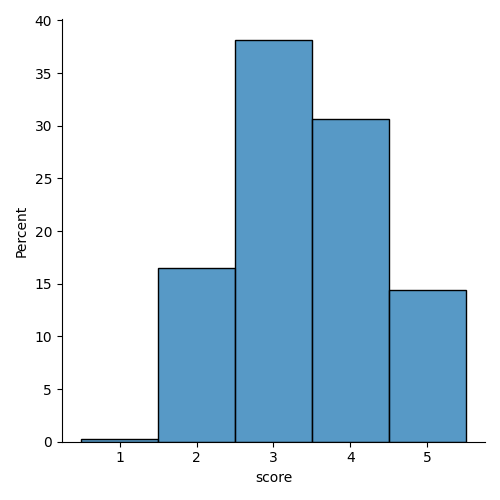}
        \caption{In-Context Examples}
        \label{fig:in-context_examples}
    \end{subfigure}
    \begin{subfigure}[h]{0.45\textwidth}
        \centering 
        \includegraphics[width=\textwidth]{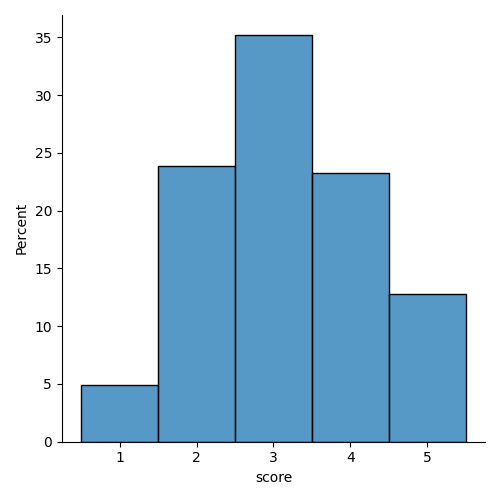}
        \caption{Agent Scaffold}
        \label{fig:agent_scaffold}
    \end{subfigure}
    \begin{subfigure}[h]{0.45\textwidth}
        \centering 
        \includegraphics[width=\textwidth]{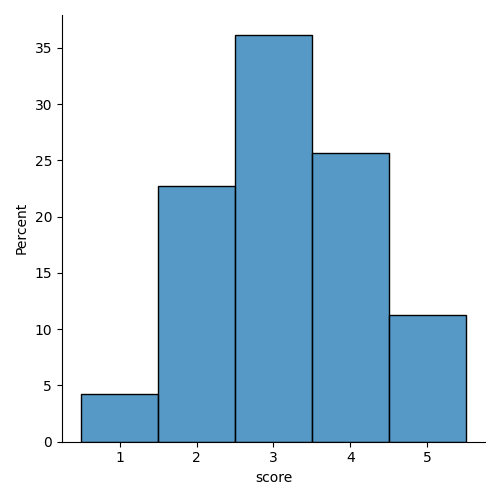}
        \caption{Distilled Skill}
        \label{fig:distilled_skill}
    \end{subfigure}
    \caption{Rubric score distributions across four generation settings for \hyphentt{Claude-Sonnet}. The distributions become increasingly bell-shaped as the augmentation level increases.}
    \label{fig:distribution_analysis}
\end{figure}

\clearpage

\section{Paper Categoristion}
\label{app:paper_category}
\vspace{2cm}
\begin{table}[h]
    \centering
    \begin{adjustbox}{max width=0.95\textwidth}
        \begin{tabular}{l}
            \toprule
            \textbf{Paper Title} \\
            \midrule
            \textbf{A. Language Model Adaptation and Alignment}\\
            \midrule
            \textit{APT: Adaptive Pruning and Tuning Pretrained Language Models for Efficient Training and Inference} \\
            \textit{A Mechanistic Understanding of Alignment Algorithms: A Case Study on DPO and Toxicity} \\
            \textit{BBox-Adapter: Lightweight Adapting for Black-Box Large Language Models} \\
            \textit{Stay on Topic with Classifier-Free Guidance} \\ 
            \textit{What Will My Model Forget? Forecasting Forgotten Examples in Language Model Refinement} \\
            \midrule
            \textbf{B. Vision and Vision-Language Models} \\
            \midrule
            \textit{Bridging Data Gaps in Diffusion Models with Adversarial Noise-Based Transfer Learning}\\
            \textit{LCA-on-the-Line: Benchmarking Out-of-Distribution Generalization with Class Taxonomies} \\
            \textit{Robust CLIP: Unsupervised Adversarial Fine-Tuning of Vision Embeddings for Robust Large Vision-Language Models} \\
            \textit{Sample-specific Masks for Visual Reprogramming-based Prompting} \\
            \textit{Test-Time Model Adaptation with Only Forward Passes} \\
            \midrule
            \textbf{C. Reinforcement Learning} \\
            \midrule
            \textit{Fine-tuning Reinforcement Learning Models is Secretly a Forgetting Mitigation Problem} \\
            \textit{RICE: Breaking Through the Training Bottlenecks of Reinforcement Learning with Explanation} \\
            \textit{SAPG: Split and Aggregate Policy Gradients} \\
            \textit{Unsupervised Zero-Shot Reinforcement Learning via Functional Reward Encodings} \\
            \midrule
            \textbf{D. Probabilistic and Mathematical ML} \\
            \midrule
            \textit{All-in-one Simulation-Based Inference} \\
            \textit{Batch and Match: Black-Box Variational Inference with a Score-Based Divergence} \\
            \textit{Challenges in Training PINNs: A Loss Landscape Perspective} \\
            \textit{Refined Coreset Selection: Towards Minimal Coreset Size under Model Performance Constraints} \\
            \textit{Sequential Neural Score Estimation: Likelihood-Free Inference with Conditional Score Based Diffusion Models} \\
            \textit{Stochastic Interpolants with Data-Dependent Couplings} \\
            \bottomrule

        \end{tabular}
        \end{adjustbox}
       
    \caption{Paper categorisation for PaperBench}
    \label{tab:qualitative_analysis}
    \vspace{-1em}
\end{table}
\clearpage
\twocolumn
\section{Threshold Discussion}
\label{app:threshold_discussion}

We report the main intrinsic evaluation results with the matching threshold set to $\tau=0.7$ in Table~\ref{tab:intrinsic_results}. To assess sensitivity to this choice, we also provide results for $\tau=0.65$ and $\tau=0.75$. Across these threshold values, our key finding remains consistent: stronger generation settings primarily improve recall, while precision fluctuates across settings.
\begin{table}[h]
    \centering
    \small
    \begin{adjustbox}{max width=0.95\columnwidth}
    \begin{tabular}{lcccc}
    \toprule
    \multirow{2}{*}{\textbf{Method}} & \multirow{2}{*}{\textbf{Backbone Model}} & \multicolumn{3}{c}{\textbf{Cosine Similarity}}\\
    \cmidrule(lr){3-5}
    && \textbf{Precision} & \textbf{Recall} & \textbf{F1}\\
    \midrule
    Human Annotation & - & 0.62 & 0.42 & 0.49 \\
    \midrule
    Direct Prompting & \multirow{4}{*}{\hyphentt{Claude-Sonnet}} & \textbf{0.57} & 0.31 & 0.38  \\
    ~\textit{+In-Context Examples} & & 0.53 & 0.36 & 0.41 \\
    ~\textit{+Agentic Scaffold} & & 0.47 & 0.36 & 0.39 \\
    ~\textit{+Distilled Skill} & & 0.54 & \textbf{0.41} & \textbf{0.44} \\
    \midrule
    Direct Prompting & \multirow{4}{*}{\hyphentt{GPT-5.4}} & 0.41 & 0.24 & 0.28 \\
    ~\textit{+In-Context Examples} & & 0.42 & 0.29 & 0.33 \\
    ~\textit{+Agentic Scaffold} & & \textbf{0.49} & 0.32 & \textbf{0.37} \\
    ~\textit{+Distilled Skill} & & 0.43 & \textbf{0.35} & \textbf{0.37}  \\
    \bottomrule
\end{tabular}
    \end{adjustbox}
    \vspace{-0.5em}
    \caption{Intrinsic meta-evaluation results with $\tau=0.75$}
    \label{tab:cos_sim_0.75}
\end{table}

\begin{table}[h]
    \centering
    \small
    \begin{adjustbox}{max width=0.95\columnwidth}
    \begin{tabular}{lcccc}
    \toprule
    \multirow{2}{*}{\textbf{Method}} & \multirow{2}{*}{\textbf{Backbone Model}} & \multicolumn{3}{c}{\textbf{Cosine Similarity}}\\
    \cmidrule(lr){3-5}
    && \textbf{Precision} & \textbf{Recall} & \textbf{F1}\\
    \midrule
    Human Annotation & - & 0.91 & 0.85 & 0.88 \\
    \midrule
    Direct Prompting & \multirow{4}{*}{\hyphentt{Claude-Sonnet}} & \textbf{0.90} & 0.78 & 0.83  \\
    ~\textit{+In-Context Examples} & & 0.88 & 0.80 & 0.83 \\
    ~\textit{+Agentic Scaffold} & & 0.87 & 0.83 & 0.84 \\
    ~\textit{+Distilled Skill} & & 0.88 & \textbf{0.86} & \textbf{0.87} \\
    \midrule
    Direct Prompting & \multirow{4}{*}{\hyphentt{GPT-5.4}} & 0.87 & 0.77 & 0.81 \\
    ~\textit{+In-Context Examples} & & 0.87 & 0.78 & 0.82\\
    ~\textit{+Agentic Scaffold} & & \textbf{0.88} & 0.79 & 0.82 \\
    ~\textit{+Distilled Skill} & & 0.84 & \textbf{0.81} & \textbf{0.83} \\
    \bottomrule
\end{tabular}
    \end{adjustbox}
    \vspace{-0.5em}
    \caption{Intrinsic meta-evaluation results with $\tau=0.65$}
    \label{tab:cos_sim_0.65}
\end{table}

\clearpage
\end{document}